\documentclass[runningheads]{llncs}
\usepackage{graphicx}
\usepackage{amsmath}
\usepackage{hyperref}
\usepackage{url}
\usepackage{acronym}
\usepackage{todonotes}
\usepackage{booktabs}
\acrodef{ET}{enhancing tumor}
\acrodef{TC}{tumor core}
\acrodef{WT}{whole tumor}

\acrodef{ML}{Machine Learning}
\acrodef{FL}{Federated Learning}

\acrodef{DSC}{Dice similarity score}
\acrodef{HD}{Hausdroff distance}

\acrodef{SGD}{Stochastic Gradient Descent}
\acrodef{RegAgg}{Regularised Aggregation Policy}

\acrodef{FeTS}{Federated Tumor Segmentation Challenge}

\begin{document}
\title{Robust Learning Protocol for 
Federated Tumor Segmentation Challenge}
%
%
\author{Ambrish Rawat\inst{1} \and
Giulio Zizzo\inst{1}\and
Swanand Kadhe\inst{2}\and
Jonathan P.\ Epperlein\inst{1}\and
Stefano Braghin\inst{1}}
\authorrunning{A. Rawat et al.}
%

\institute{IBM Research Europe, Dublin, Ireland\\%
\email{\{ambrish.rawat@ie, giulio.zizzo2, jpepperlein@ie, stefanob@ie\}.ibm.com} \and
IBM Research, Almaden, USA\\%
\email{swanand.kadhe@ibm.com}}

\maketitle              

\begin{abstract}
In this work, we devise robust and efficient learning protocols for orchestrating a \ac{FL} process for the Federated Tumor Segmentation Challenge (FeTS 2022). Enabling FL for FeTS setup is challenging mainly due to data heterogeneity among collaborators and communication cost of training. To tackle these challenges, we propose a \textbf{Ro}bust \textbf{Le}arning \textbf{Pro}tocol (RoLePRO) which is a combination of server-side adaptive optimisation (e.g., server-side Adam) and judicious parameter (weights) aggregation schemes (e.g., adaptive weighted aggregation). RoLePRO takes a two-phase approach, where the first phase consists of vanilla Federated Averaging, while the second phase consists of a judicious aggregation scheme that uses a sophisticated re-weighting, all in the presence of an adaptive optimisation algorithm at the server. We draw insights from extensive experimentation to tune learning rates for the two phases.  
\end{abstract}
\section{Introduction}

In this work we investigate a federated learning system for Brain Tumor Segmentation as presented in the \ac{FeTS}~\cite{Pati2021_FeTS,Reina2021_OpenFL,Baid2021_BraTS}.
Brain Tumor Segmentation is a medical imaging task where given a an MRI scan a model is tasked to produce the segmentation demarcating the regions corresponding to a tumor namely, \ac{ET}, \ac{TC} and \ac{WT}.
Traditionally, a deep learning model with a U-Net architecture~\cite{Ronneberger2015_UNet} is trained for this task.
The data for training such models often resides with different institutions which prohibits the use of classical \ac{ML} pipelines that require the data to be available centrally at one location.
Therefore, in such situations one may resort to the privacy-preserving paradigm of \ac{FL}~\cite{McMahan2017_Communication} for training the model, which is the focus of \ac{FeTS}.
This challenge presents two tasks - the first involves algorithms for \ac{FL} orchestration for improved model convergence, and the second is testing the generalisability of a model in the ``wild'' on the data of clients who did not participate in the original federation.  
We primarily focus on the first task and investigate a combination of approaches for client- and server-side optimisation schedules, parameter aggregation and fine tuning to help with model convergence.

\section{\ac{FeTS} Setup}
\label{sec:fets_setup}

The federation consists of 23 different institutions~\cite{Pati2021_FeTS,Reina2021_OpenFL,Baid2021_BraTS} seeking to collaboratively learn a U-Net model~\cite{Ronneberger2015_UNet} for brain tumor segmentation.
Each participating institution owns a variably sized data partition which resides privately with the host.
The details of the partitioning are specified in \texttt{partitioning1.csv} where one can notice a skewed distribution with two collaborators, namely  and  together hold a major chunk of the total data.
On top of this, the classical iid assumption of machine learning often goes for a toss and the respective partitions may have highly non-iid characteristics.
Such skewed distributions are not foreign for federated setups and often the learning protocols are appropriately modified to account for the heterogeneity.
In order to moderate the heterogeneity, the challenge also presents an additional partitioning where some institutions are further split based on the tumour size, resulting in a more balanced distribution across $33$ clients as specified in \texttt{partitioning2.csv}.

More generally, such cross-silo \ac{FL} setups consists of $M$ clients, where $m\textsuperscript{th}$ client owns data $D_m = \{(x_i,y_i)\}_{i=1}^{n_m}$ with $n_m$ samples.
The training is performed iteratively across multiple \ac{FL} rounds.
In round $t$, first the central server or aggregator selects a set $C_{(t)}$ of clients, referred to as a collaboration, for training.
Second it broadcasts the current set of parameter vector W =$\{w_{t,j}\}$ to all $N$ clients for computing a set of validation metrics (for brevity, we will often drop the dependence on $j$ and refer to a single parameter as $w$).
Third the clients in the collaboration $C_{(t)}$ train the model on their local data partitions to obtain the updated parameter $w_{t+1}^c$ along with a set of post-training validation metrics, both of which they share with the aggregator.
Finally the aggregator combines the weight vectors from different clients typically with weighted averaging (FedAvg) as, $w_{t+1} = \sum_{c\in C_{(t)}}  n_c {w_{t+1}^c}/ N_{C_{(t)}}$ where $N_{C_{(t)}} = \sum_{c\in C_{(t)}} n_c$ is the sum of samples across clients in collaboration $C_{(t)}$. The federation is performed for a total of $T$ rounds with suitable measures like early stopping to help reduce the generalisation error.\looseness=-1

\paragraph{\bfseries Metrics.} There are two metrics tracked as part of \ac{FeTS} setup - \ac{DSC} which measures the overlap between predicted and ground truth segmentations and \ac{HD} which accounts for the distance between segmentation boundaries. The details for these metrics are described in~\cite{Pati2021_FeTS}. During the \ac{FL} training, \ac{FeTS} maintains the checkpoint of model which achieves the best \ac{DSC} on the validation set which itself is set locally by each client with a classical 80-20 split.

\paragraph{\bfseries Modelling Challenges.} There are two key challenges in enabling federated learning for such settings. The first stems from the data distributions across different clients which are typically not iid. This often leads to diverging updates during training rounds which hamper model convergence. And second is the communication cost of training rounds and the presence of stragglers in the federation, both of which add an overhead for the learning process.

Typically \ac{FL} setups either belong to a cross-silo or a cross-device setting. The former comprises of a clientele of the order of 100 participants each holding a relatively large data set while the latter could have as many as a billion clients with only few data samples associated with each participant. A synchronous orchestration with all clients participating in every round adds a massive communication overhead which can adversarially affect the rate of convergence. On the other hand, ignoring certain clients could bias the resultant model towards specific modes thereby deteriorating their generalisation capability.

\paragraph{\bfseries Practical Challenges.} While \ac{FeTS} presents an interesting setup for exploring a combination of schemes for \ac{FL} orchestration, it is worth noting that there are inherent assumptions in the system. An understanding of these assumptions helps in scoping out the playing field and shaping strategies for algorithm design and experimentation. For instance, the APIs for Task 1 limit access to local training protocols of participating clients which limits the the applicability of approaches like FedProx~\cite{Li2020_FedProx} and Scaffold~\cite{Karimireddy2020_Scaffold} to tackle the non-iidness in the system. Similarly, the available deployment of \ac{FeTS} requires relatively large compute, of the order of 300~GB of CPU RAM optionally with at least 16~GB of GPU memory. Additionally, each training rounds can take up to 7 hours which can delay the feedback often required for rapid prototyping, hyperparameter optimisation and prohibit re-runs for marginalising the experimentation noise. Finally, it should be mentioned that the orchestration maintains a counter for total simulated time which accounts for the different costs in a typical FL process including communication cost, training time and cost related to computation of validation metrics. The script exits when the total simulated time reaches one week. It is also worth mentioning that the specified FL plan for \ac{FeTS} involves global computation of metrics, i.e.\ each client, irrespective of their participation in the collaboration, computes the set of validation metrics on the latest aggregated model. This, as we discuss in Section~\ref{sec:approach}, can be beneficial for devising both client selection and parameter aggregation strategies. In summary, we work within these constraints and devise schemes that can help improve the rate of model convergence within the permissible simulated time.

\paragraph{\bfseries \ac{FeTS} 2021.} The challenge was also hosted in 2021~\cite{Pati2021_FeTS} with some notable differences. To our understanding this year's challenge includes data from additional institutions potentially exacerbating the non-iidness in the system. We also note that Cost Weighted Averaging~\cite{Machler2021_FedCostWAgg} and Adaptive Weight Averaging~\cite{Khan2021_Adaptive2} resulted in winning solutions for the posed challenge which we explore within our overall FL plan as described in Section~\ref{sec:approach}.




\section{Our Approach}
\label{sec:approach}

\begin{figure}
    \centering
    \includegraphics[width=0.85\textwidth]{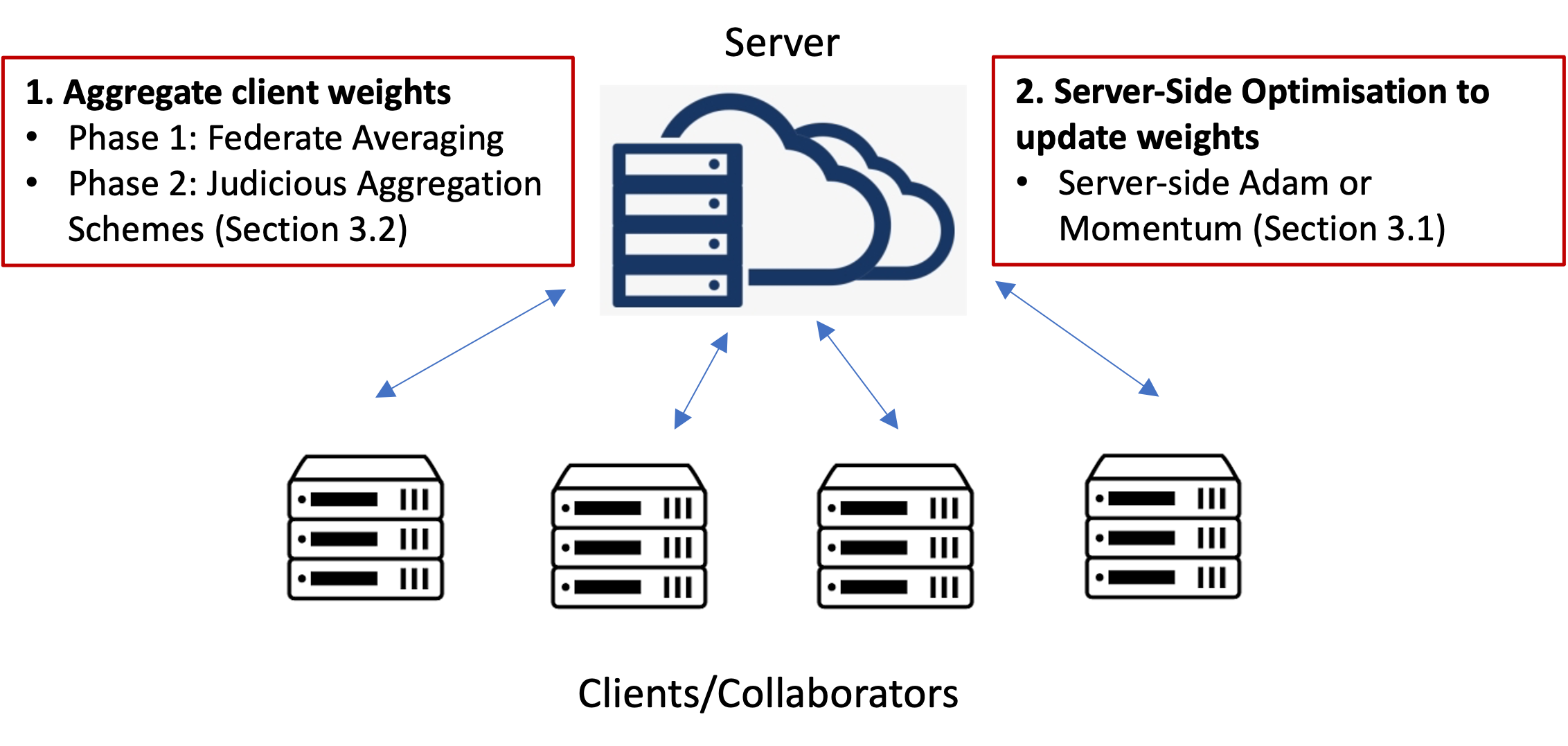}
    \caption{Summary of our approach: a combination of server-side adaptive optimisation and judicious parameter aggregation in a two-phase process with all clients participating in every round.}
    \label{fig:approach}
\end{figure}

\paragraph{\bfseries Summary of our approach.} 
We propose to use a combination of server-side adaptive optimisation and judicious parameter aggregation in a two-phase process with all clients participating in every round. In the first phase, the server aggregates the weights with vanilla FedAvg and in the second it employs a judicious aggregation scheme which uses a more sophisticated re-weighting (we discuss several judicious aggregation schemes later). In both phases, server-side adaptive optimisation (e.g., server-side Adam described later) is used. The learning rates are appropriately adjusted for the two phases. We summarise our approach in Fig.~\ref{fig:approach}. In the remaining section, we describe our thinking behind the proposed approach and provide details of its various components.

\ac{FeTS} presents a highly constrained and challenging scenario where both modelling and experimentation challenges need to be addressed simultaneously. These constraints limit the use of brute-force approaches for algorithm design, experimentation, and validation as well as hyperparameter optimisation. We therefore break the overall task into smaller components with a seat-of-the-pants strategy. First, we run the vanilla FedAvg algorithm with the default settings and make some observations. This involves every client participating in every round of the collaboration. Based on these observations we conjecture a working theory for this setup and use it to devise experiments. In particular, we note the behaviour of the optimisation and adopt a suitable learning rate schedule to fit this scenario. Additionally, we also adopt the server-side optimisation with adaptive schemes to help speed up convergence. We then look at a set of different aggregation schemes which are inspired from the successful solutions of \ac{FeTS} 2021. Finally, we consider a two-phase process where we consider aggregation schemes as part of a fine tuning strategy. In this approach we first obtain a reasonable initialisation within first few FL rounds with vanilla FedAvg and follow it up with \textit{fine-tuning} phase for the remaining time. For all of these experiments we only adopt the strategies for weights and biases in the U-Net. All other parameters are aggregated with weights proportional to sample size. It is worth noting that, even though we primarily focus on all clients participating in every round, our approach can in general be adapted to include client selection.


\paragraph{\bfseries Default Setup.} We first run the experiment with the default setting which consist of: all clients participating in every round, vanilla federated averaging, and fixed hyperparameters of learning rate and epochs per round across all rounds. We note the following observations from this experiment - 1) while there are two different partitions available we note that \texttt{partitioning2.csv} naturally provides a better fit with data parallelism, 2) the optimisation is not stable with an oscillating objective during the optimisation, 3) we experiment with local epochs of 1 and 3 and do not note any remarkable difference in DICE scores. We also note that it takes an average of 700 minutes (of simulated time) per round with all clients participating. Thus, with all 33 clients training for one local epoch in each round, one can perform a federation for up to 16 training rounds within the 1 week limit specified in the setup. 


\subsection{Server-Side Optimisation}
\label{sec:optalgo}

In order to dampen the oscillations we explore the use of adaptive optimisation strategies. An \ac{FL} setup consists of two different 
rates - a client side learning rate $\lambda_c$ and a server side learning rate $\lambda_s$. In general, at round $t$ the server first computes an aggregate of obtained updates $\hat{w}_{t+1}$ from the clients using the specified aggregation scheme, which is used to form a proxy $\Delta_{t} = w_t - \hat{w}_{t+1}$ for the gradient from the server's perspective. It then uses an optimisation strategy to modify its parameter state with the obtained delta. Thus, vanilla FedAvg can be thought of as performing a server side \ac{SGD} $w_{t+1} = w_{t} - \lambda_s\Delta_t$ with $\lambda_s=1.0$. However, adaptive strategies like momentum-based aggregation or even server-side Adam can be used to accelerate the convergence and dampen the oscillations~\cite{Reddi2021_Adaptive,Wang2021_Field}. It is worth pointing out that in \ac{FeTS} the client-side optimisation is fixed as Adam~\cite{Kingma2015_Adam}. The works of~\cite{Reddi2021_Adaptive} and ~\cite{Wang2021_Field} provide an in-depth analysis of optimisation schemes for federated setup with some useful insights. We focus on two key takeaways from this work - they recommend jointly tuning $\lambda_c$ and $\lambda_s$ as they observe that the accuracy often follows a staircase pattern for adaptive schemes, and to decay the client-side learning rates when clients take more than a few gradient steps in their local optimisation. Finally, they also note that momentum-based adaption is more sensitive to choice of learning rates, while FedAdam and FedYogi~\cite{Reddi2021_Adaptive} are more stable with respect to these choices.

For this experiment we stick with the classical sample-size-based weighting for aggregation and explore the use of adaptive optimisation strategies at the server end. We later adopt the learning from this experiment across other aggregation schemes. For convenience we refer to this intermediate parameter state as $\hat{w}_t$ which refers to the parameter obtained after using an aggregation scheme at the server end. This is subsequently used in combination with different optimisation approaches\looseness=-1

\begin{itemize}
    \item \textbf{OptAlgo A}: server-side \ac{SGD} with $\lambda_s=1.0$ combined with step-wise constant learning rates for clients across different rounds where $\lambda_c=5\cdot10^{-5}$ for first 7 rounds followed by $\lambda_c=5\cdot10^{-6}$ for the remaining rounds.
    \vspace{2mm}
    
    \item \textbf{OptAlgo B}: server-side momentum where a moving average is maintained with respect to change in $\Delta_t$ which is in turn used to update the parameter state. The update can be summarised as,
    
    \begin{equation}
        \begin{aligned}
            \Delta_{t} &= w_t - \hat{w}_{t+1}\\
            m_{t} &=\beta m_{t-1}+ \Delta_{t} \\
            w_{t+1} &=w_{t}-\lambda_s m_{t}
        \end{aligned}
    \end{equation}
    This has been explored in~\cite{Isik2021_Federated} and an implementation is available in \ac{FeTS}. We use a $\beta$ of 0.9 for this approach and fix $\lambda_s$ as 0.1 and $\lambda_c=5\cdot10^{-4}$ for first 7 rounds followed by $\lambda_c=5\cdot10^{-5}$ for the remaining rounds.
    \vspace{2mm}
    
    \item \textbf{OptAlgo C}: server-side Adam adopts a parameter-specific update by upweighting updates for parameters which receive sparse updates
    
    \begin{equation}
        \begin{aligned}
            \Delta_{t} &= w_t - \hat{w}_{t+1} \\
            m_{t} &=\beta_{1} m_{t-1}+\left(1-\beta_{1}\right) \Delta_{t} \\
            v_{t} &=\beta_{2} v_{t-1}+\left(1-\beta_{2}\right) \Delta_{t}^{2}  \\
            w_{t+1} &=w_{t}-\lambda_s \frac{m_{t}}{\sqrt{v_{t}}+\tau}
        \end{aligned}
    \end{equation}
    
    FedAdam has previously been investigated in~\cite{Reddi2021_Adaptive} where the authors fix $\tau$ as $0.001$ and $\beta_1$ and $\beta_2$ as 0.9 and 0.99 respectively. They further experiment with a range of values for $\lambda_c$ and $\lambda_s$ and note that it often follows a staircase pattern. Following some initial experiments we note that $\lambda_s$ of 0.001 and $\lambda_c$ of $5\cdot10^{-4}$ provides stable updates. While this scheme differs slightly from the original Adam implementation where the $\tau$ is added as part of the square root and $m_t$ and $v_t$ are scaled with a decay for every new update, we didn't find an empirical impact on the algorithm when ran for few rounds. 

\end{itemize}

\begin{figure}
    \centering
    \includegraphics[width=0.49\textwidth]{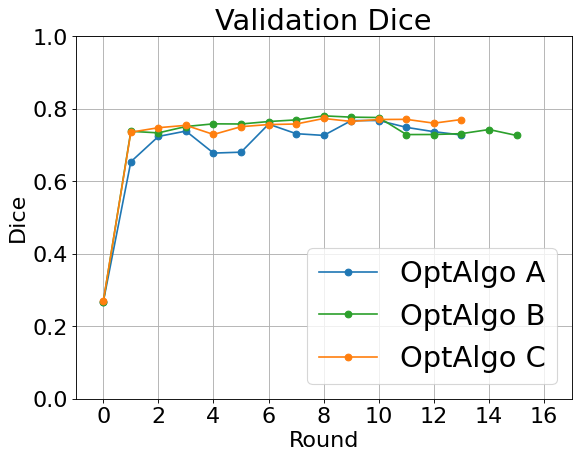}
    \includegraphics[width=0.49\textwidth]{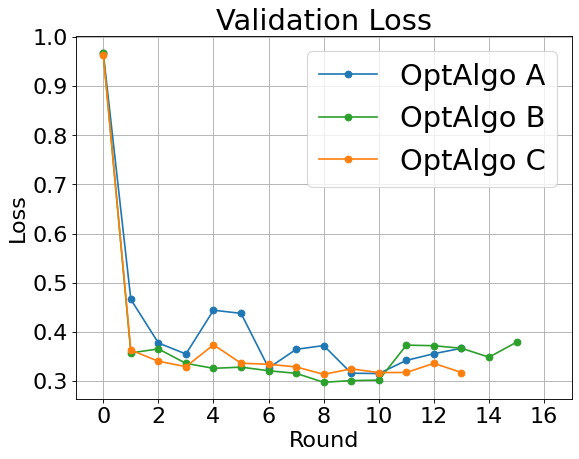}
    \caption{Performance of Optimisation Algorithms with FedAvg for weight aggregation.}
    \label{fig:optalgos}
\end{figure}

It was clear in our experimentation and as is shown in Figure~\ref{fig:optalgos} that adaptive schemes of OptAlgo B and OptAlgo C help dampen the oscillations observed for OptAlgo A. We would like to emphasise that we experimented in an ad-hoc fashion with manual intervention at different stages. For example, in some experiment runs that lower the $\lambda_c$ further by a factor of 10 also helped with convergence. We suspect that normalising the number of updates across clients or reducing the local epochs across rounds with a decay factor can further help with the convergence. In general, our observations were consistent with those made in~\cite{Reddi2021_Adaptive} and~\cite{Wang2021_Field} - both OptAlgo B and OptAlgo C improved convergence, and OptAlgo C was the most stable with respect to choice of $\lambda_c$ resulting in similar convergence behaviour across a range of different values for $\lambda_c$ and $\lambda_s$. However, the absence of rigorous experimentation with hyperparameters can potentially result in an apples-vs-oranges comparison. We therefore take the teachings with a pinch of salt and only draw some high level conclusions as in the absence of rigorous testing they only serve as guiding approaches for our subsequent experiments. 

\subsection{Parameter Aggregation Methods}
\label{sec:param_agg}

Inspired from the approaches in \ac{FeTS} 2021, we also experiment with a few different parameter aggregation schemes. These refer to the set of algorithms that are used to obtain $\hat{w}_{t+1}$ from a set of client updates $\{w^c_t\}_{c\in C_{(t)}}$ received in round $t$. We replicate the setup of~\cite{Machler2021_FedCostWAgg} and~~\cite{Khan2021_Adaptive2} to the best of our understanding and experiment with some modifications which are described below.  

\begin{itemize}\itemsep2ex
    \item \textbf{CostWAgg} : Cost Weighted Aggregation (CostWAgg)~\cite{Machler2021_FedCostWAgg} uses the change in the validation loss across different rounds to guide the weighting during parameter aggregation. More specifically, for each client $c \in C_{(t)}$ it first computes the normalised ratio $r_c =l_c(w^c_{t-1})/l_c(w^c_{t})$ of the local validation loss (after training) from the previous round $l_c(w^c_{t-1})$ with the one obtained for the current round $l_c(w^c_{t})$, thereby giving larger weight to updates which resulted in larger decrease in loss. They further take a convex combination of the cost-based weight and sample-size-based weight to obtain the final parameter vector. In their scheme all clients participate in every round, i.e.\ $C_{(t)} = \{1,\dotsc,M\}$ $\forall t$, and train for 10 epochs locally before sharing the updates. The update in round $t$ is obtained as
    \begin{equation}
        \begin{aligned}
            \hat{w}_{t+1} &=\sum_{c\in C_{(t)}}\left(\alpha \frac{n_{c}}{N_{C_{(t)}}}+(1-\alpha) \frac{r_{c}}{R_{C_{(t)}}}\right) w_{t}^{c},
        \end{aligned}
    \end{equation}
    where $R_{C_{(t)}} = \sum_{c\in C_{(t)}} r_c$ and $\alpha$ is a hyperparameter for balancing the two terms. The authors recommend  $\alpha = 0.5$ while remarking that the optimisation wasn't particularly sensitive to the choice of $\alpha$. Also, note that this scheme results in identical weighting across all parameters. We use a variant of this scheme where each client computes the ratio with respect to the local validation loss computed before and after training in the \emph{current} round, and refer to it as \textbf{RoundCWAgg}. This ratio can be computed as $l_c(w_{t-1})/l_c(w^c_{t})$. Note that while CostWAgg requires the use of $l_c(w^c_{t-1})$ which is only available for clients which participated in the previous training round, RoundCWAgg doesn't require the previous loss for the current collaboration, since $l_c(w_{t-1})$ is available for all clients in the federation. We use $\alpha=0.1$ to upweight the loss based contribution for RoundCWAgg.
    
    \item \textbf{\acs{RegAgg}:} Adaptive Weighted Aggregation Policy~\cite{Khan2021_Adaptive2} proposes a parameter weighting scheme that upweights contributions that are close to the average update. First, for each client $c$ it computes an inverse of the absolute difference between client parameter $w_t^c$ and the mean of the collaboration. These are normalised across clients to obtain the client-specific weighting factor $u^w_c$ for the parameter $w$ where clients whose updates are close to the average are assigned larger weights. This is then used in combination with sample-size-based weights $\nu_c = n_c/N_{C_{(t)}}$ to form the final weights with either an additive operator $u^w_c + \nu_c$ or a multiplicative operator $u^w_c \cdot \nu_c$. The former is referred to as Similarity Aggregation Policy (SimAgg) and the latter as \ac{RegAgg} Policy. The authors use random sampling for selecting clients for collaboration (while ensuring fair selection of all clients) and remark that \ac{RegAgg} with 5 local epochs provided them the best results during their experimentation. \ac{RegAgg} updates every  scalar weight in the weight vector $\hat{w}_{t+1}$ individually. We use $\omega=\hat{w}_{t+1}$ to reduce clutter.
    \begin{equation}
        \begin{aligned}
            \omega_{\text{mean}} &= \frac{1}{\# C_{(t)}} \sum_{c \in C_{(t)}}w_{t}^c\\
            u^w_{c} &= \frac{1/\left| w^c_{t} - \omega_{\text{mean}}\right|_\tau}{\sum_{i\in C_{(t)}}1/\left| w^i_{t,j} - \omega_{\text{mean}}\right|_\tau}\\
            \omega&=\frac{\sum_{c \in C_{(t)}}\left(u^w_c\cdot \nu_c \cdot w^c_{t}\right)}{\sum_{c \in C_{(t)}}\left(u^w_{c}\cdot \nu_c\right)},
        \end{aligned}
    \end{equation}
    where $|x|_\epsilon:=|x|+\epsilon$ avoids divisions by zero; we use $\epsilon=1\cdot10^{-5}$ as specified in~\cite{Khan2021_Adaptive2}.
    This aggregation scheme results in different weights for different parameters, since $u^w_c$ for every parameter $w$. We also experiment with a modification to this algorithm \textbf{RegMedAgg} which uses median ($\omega_{\text{median}}$) for computing $u^w_c$ as opposed to the arithmetic mean.
    
    \item \textbf{RegCostAgg}: This approach combines the approaches of CostWAgg and \ac{RegAgg} where the loss ratio $r_c$ is combined with the sample-size weight in a multiplicative way. This results in,
    \begin{equation}
        \begin{aligned}
            \hat{w}_{t+1}&=\frac{\sum_{c \in C_{(t)}}\left(r_c\cdot \nu_c \cdot w^c_{t}\right)}{\sum_{c \in C_{(t)}}\left(r_c\cdot \nu_c\right)}
        \end{aligned}
    \end{equation}
\end{itemize}

We run 13 rounds of FL training and use OptAlgoB while comparing these different parameter aggregation scheme. We specifically monitor four metrics - best validation Dice across all labels till round 13, best validation loss till round 13 as well as average validation DICE and validation loss for last five rounds. The results are summarised in Table~\ref{tab:param_agg}. While in terms of Best Dice (or loss) values many methods are comparable, the Avg Dice (or loss) tells a more relevant story as it sheds light on the stability of the overall optimisation protocol. Mindful of the fact that these numbers weren't obtained over multiple runs, we tentatively conclude that the parameter aggregation schemes provide marginal benefits over vanilla FedAvg (with generally smaller avg loss and higher avg Dice).

\begin{table}[h]
  \centering
      \caption{Performance of different parameter aggregation schemes with OptAlgoB for optimisation. Average values show the stability of the FL process near convergence.}
 \label{tab:param_agg}
  \begin{tabular}{l|cccc}
    \hline\noalign{\smallskip}
    \multicolumn{1}{l}{} & Best Dice  & Avg Dice & Best loss & Avg loss\\
    \noalign{\smallskip}
    \hline
    \noalign{\smallskip}
    FedAvg      &  0.7771   & 0.7417    & 0.3008    & 0.3522\\
    \midrule
    CostWAgg    & 0.7738    & 0.7578    & 0.2984    & 0.3157\\
    RoundCWAgg  & 0.7718    & 0.7022    & 0.3074    & 0.4095\\
    \midrule
    RegAgg      & 0.7783    & 0.7310    & 0.2920    & 0.3608\\
    RegMedAgg   & 0.7569    & 0.7448    & 0.3254    & 0.3478\\
    \midrule
    RegCostAgg  & 0.7848    & 0.7595    & 0.2900    & 0.3272 \\
    \noalign{\smallskip}
    \hline    
  \end{tabular}
\end{table}

\subsection{Fine Tuning Approaches}

In Section~\ref{sec:optalgo} we noted that OptAlgoB and OptAlgoC help dampen the oscillations and stabilise the server-side optimisation, and in Section~\ref{sec:param_agg} we noted some benefits in the use of different parameter aggregation schemes over FedAvg. With these two lessons and the proverbial knowledge among practitioners that the benefits of domain specific adaption are more suitable as fine tuning, we conjecture an FL plan - \textbf{Ro}bust \textbf{Le}arning \textbf{Pro}tocol (RoLePRO) that consists of two stages: First, we run vanilla FedAvg for a few initial rounds with all clients participating in each round with either OptAlgoB or OptAlgoC and appropriate choices of learning rates $\lambda_s$ and $\lambda_c$. Then, we fine-tune the obtained model with the one of the following aggregation schemes - CostWAgg, RegAgg, TrimmedMean and TopKRegCost, and preferably a reduced set of learning rates. The use of TrimmedMean and TopKRegCost is motivated by an intent to reduce the effect of variance within client updates during aggregation which we explain below.\looseness=-1

As described in Section~\ref{sec:fets_setup} the data split across the different participating institutions is highly imbalanced. Naturally, such scenarios lead to large variance among the client updates in each round. It is worth remarking that within FeTS the institutions share updates after a fixed number of epochs which is common for all clients. Thus, clients perform a varying number of local gradient updates which can exacerbate client drift and result in disparate client updates. The success of CostWAgg and RegAgg in \ac{FeTS} also bears a mention here. CostWAgg effectively prefers updates with more stable loss changes across different rounds and combines it with sample weights thereby giving a high weight to smoothed updates from large-data clients. Similarly, RegAgg enables preferential weighting by exploiting the geometry of updates in combination with the sample weighting. Effectively, updates which are close the mean-update and correspond to clients with large data receive the largest weights during aggregation. Such weighting for \ac{FeTS} has also been explored in~\cite{Luo2021_Rethinking} where they categorised the participating institutions as internal and external depending on their sample contribution and developed different weighting schemes for the two groups.
TrimmedMean provides an alternative way to alleviate the variance by only accounting for the geometry where first the set of updates are trimmed by discounting the ones that are too far from the median, and the remaining fraction of filtered updates are averaged (without any weighting). This has been explored in the context of byzantine behaviour in Federated Learning~\cite{Mhamdi2018_Hidden,Baruch2019_Alie}. While this approach can be beneficial for softening the aggregation, it often slows down convergence. Also worth noting that for TrimmedMean, the filtering of updates across different clients before the aggregation step is different for different parameters. In TopKRegCost we account for loss values and sample size for this filtering step by first sorting the clients as per their loss-ratio and sample contribution with the score, $n_c/N_{C(t)}\cdot l_c(w^c_{t-1})/l_c(w^c_{t})$, similar to RegCostAgg, and then averaging (without weighting) the $k$ best updates with the largest scores. We use a filtering factor of 20\% for both TrimmedMean and TopKRegCost. Note that TopKRegCost can be thought of as an aggressive or harder version of its soft counterparts RegCostAgg and CostWAgg where instead of down-weighting fruitless contributors one simply discards them and is democratic in its consideration of filtered contributors. Contrary to TrimmedMean, the filtering in TopKRegCost is common across all parameters during a training round. 

All of the aforementioned aggregation schemes can result in biased models. For instance, TopKRegCost can lead to a preferential treatment of a select few clients, especially with a high filtering factor. The hyperparameter $\alpha$ controls a similar trade-off for CostWAgg. Such loss-based aggregation has also been explored in other contexts for FL. One example is the Federated Adversarial Training~\cite{Zizzo2020_FAT} protocol which suffers from highly disparate updates during the training rounds and modifying the weighting scheme helped improve convergence~\cite{Zhu2021_alpha}. Similarly, best-$k$ sparsification has been used to combat model poisoning in FL~\cite{Panda2022_SpaseFed}.\looseness=-1

In Table~\ref{tab:tuning_optalgob} and~\ref{tab:param_tuning_optalgoc} we present the results for RoLePro with OptAlgoB and OptAlgoC across different aggregation schemes. We observe that both RegAgg and TopKRegCost lead to high-performing models across both OptAlgoB and OptAlgoC. It should be mentioned that apart from aggregation schemes, the optimisation protocols of OptAlgoB and OptAlgoC also enable parameter specific learning rates, therefore it can be challenging to pin-point the source of the observed gains within the complex orchestration of an FL plan.

\subsection{Submission}
\begin{table}[bt]
  \centering
  \caption{Performance of different aggregation schemes with OptAlgoB in the fine-tuning phase of the two-phase learning protocol of RoLePro}
  \label{tab:tuning_optalgob}
  \begin{tabular}{l|cccc}
    \hline\noalign{\smallskip}
    \multicolumn{1}{l}{} & Best Dice  & Avg. Dice  & Best loss  & Avg. loss  \\
    \noalign{\smallskip}
    \hline
    \noalign{\smallskip}
    RegAgg      & 0.8067    & 0.7921    & 0.2639    & 0.2821 \\
    CostWAgg    & 0.7788    & 0.7681    & 0.3107    & 0.3207 \\
    TrimmedMean & 0.7928    & 0.7834    & 0.2764    & 0.2958 \\
    TopKRegCost   & 0.7965    & 0.7806    & 0.2718    & 0.2957 \\
    \noalign{\smallskip}
    \hline
  \end{tabular}
\end{table}

We used the insights from above for our final submission, which was specifically:

Rounds 1-3 used vanilla WAvg, with the Adam optimiser on both sides. The server-side learning rate was 0.003, and on the client-side it was 0.0005.

For rounds 4-16, we switched to RegAgg, with Adam optimiser on both sides and reduced learning rates of 0.002 on the server side 0.00005 on the client side. The results on the held-out FeTS 2022 test set are reported in Table~\ref{tab:test_res}.

Adam's parameters were set to the aforementioned values of $\beta_1=0.9$, $\beta_2=0.99$ and $\tau=0.001$ in all rounds and on the server and all clients.

\begin{table}[tb]
  \centering
      \caption{Performance of different aggregation schemes with OptAlgoC in the fine-tuning phase of the two-phase learning protocol of RoLePro}
 \label{tab:param_tuning_optalgoc}
  \begin{tabular}{l|cccc}
    \hline\noalign{\smallskip}
    \multicolumn{1}{l}{} & Best Dice  & Avg. Dice  & Best loss  & Avg. loss  \\
    \noalign{\smallskip}
    \hline
    \noalign{\smallskip}
    RegAgg      & 0.7958    & 0.7851    & 0.2709    & 0.2905 \\
    CostWAgg    & 0.7888    & 0.7712    & 0.2880    & 0.3060 \\
    TrimmedMean & 0.7588    & 0.7517    & 0.3218    & 0.3314 \\
    TopKRegCost   & 0.7963    & 0.7700    & 0.2799    & 0.3086 \\
    \noalign{\smallskip}
    \hline
  \end{tabular}

\end{table}


\begin{table}[tb]
  \centering
      \caption{Segmentation performance of the submitted algorithm on the FeTS 2022 test set, by label 
      (WT: Whole Tumour, TC: Tumour Core, ET: Expanding Tumour). The reported Communications Cost was 0.702.}
 \label{tab:test_res}
  \begin{tabular}{l|cc|cc}
    \hline\noalign{\smallskip}
    \multicolumn{1}{l|}{} & \multicolumn{2}{c|}{Dice} & \multicolumn{2}{c}{Hausdorff 95} \\ 
    \multicolumn{1}{l|}{}
     & Mean (Std.) & Median (Quartiles)  & Mean (Std.) & Median (Quartiles)  \\
    \hline\noalign{\smallskip}
    WT &
    0.7752 (0.1753) & 0.8316 ([0.7105,0.8975]) & 28.93 (30.02) & 14.88 ([6.428,48.51]) \\
    TC & 
    0.7785 (0.2769) & 0.9041 ([0.7609,0.9449]) & 28.81 (80.48) & 4.123 ([2.236,10.1])\\
    ET & 
    0.747 (0.2625) & 0.8492 ([0.7224,0.9114]) & 29.59 (85.79) & 2.236 ([1.414,7.45]) \\
    \noalign{\smallskip}
    \hline
  \end{tabular}

\end{table}


\section{Conclusions}

It is well known that an \ac{FL} system has many moving parts from the numerous components within its system design to the set of hyperparameters in the learning algorithm. These can often result in intractable experimentation strategies. In this work we focused on the cross-silo or enterprise setup of FeTS and developed an FL plan to orchestrate the learning over the data residing with 23 participating institutions. We achieved this by studying two different aspects of FL namely, server-side optimisation and judicious parameter aggregation schemes, which we then used to develop a robust learning protocol for the FeTS setup. This protocol prescribes a two phase process where vanilla FedAvg is followed by a fine-tuning phase that is enabled with sophisticated parameter aggregation schemes, all in the presence of an adaptive optimisation algorithm at the server end (such as server-side Adam). While we found the choice of learning rates for the two phases to be crucial for our FL plan, we acknowledge that these hyperparameter choices require thorough analysis. Finally, we hope that this empirical investigation can serve as a guiding document for tackling the underlying challenge of \ac{FeTS}.

\section*{Acknowledgements}
This work has been partially supported by the MORE project (grant agreement 957345), funded by the EU Horizon 2020 program.

\bibliographystyle{unsrt}
\bibliography{citations.bib}

\end{document}